# Semantic Similarity in a Taxonomy: An Information-Based Measure and its Application to Problems of Ambiguity in Natural Language

**Philip Resnik**                                        RESNIK@UMIACS.UMD.EDU
*Department of Linguistics and*
*Institute for Advanced Computer Studies*
*University of Maryland*
*College Park, MD 20742 USA*

## Abstract

This article presents a measure of semantic similarity in an IS-A taxonomy based on the notion of shared information content. Experimental evaluation against a benchmark set of human similarity judgments demonstrates that the measure performs better than the traditional edge-counting approach. The article presents algorithms that take advantage of taxonomic similarity in resolving syntactic and semantic ambiguity, along with experimental results demonstrating their effectiveness.

## 1. Introduction

Evaluating semantic relatedness using network representations is a problem with a long history in artificial intelligence and psychology, dating back to the spreading activation approach of Quillian (1968) and Collins and Loftus (1975). Semantic similarity represents a special case of semantic relatedness: for example, cars and gasoline would seem to be more closely related than, say, cars and bicycles, but the latter pair are certainly more similar. Rada et al. (Rada, Mili, Bicknell, & Blettner, 1989) suggest that the assessment of similarity in semantic networks can in fact be thought of as involving just taxonomic (IS-A) links, to the exclusion of other link types; that view will also be taken here, although admittedly links such as PART-OF can also be viewed as attributes that contribute to similarity (cf. Richardson, Smeaton, & Murphy, 1994; Sussna, 1993).

Although many measures of similarity are defined in the literature, they are seldom accompanied by an independent characterization of the phenomenon they are measuring, particularly when the measure is proposed in service of a computational application (e.g., similarity of documents in information retrieval, similarity of cases in case-based reasoning). Rather, the worth of a similarity measure is in its utility for the given task. In the cognitive domain, similarity is treated as a property characterized by human perception and intuition, in much the same way as notions like "plausibility" and "typicality." As such, the worth of a similarity measure is in its fidelity to human behavior, as measured by predictions of human performance on experimental tasks. The latter view underlies the work in this article, although the results presented comprise not only direct comparison with human performance but also practical application to problems in natural language processing.

A natural, time-honored way to evaluate semantic similarity in a taxonomy is to measure the distance between the nodes corresponding to the items being compared — the shorter the path from one node to another, the more similar they are. Given multiple paths, one takes the length of the shortest one (Lee, Kim, & Lee, 1993; Rada & Bicknell, 1989; Rada et al., 1989).

A widely acknowledged problem with this approach, however, is that it relies on the notion that links in the taxonomy represent uniform distances. Unfortunately, uniform link distance is difficult to define, much less to control. In real taxonomies, there is wide variability in the "distance" covered by a single taxonomic link, particularly when certain sub-taxonomies (e.g., biological categories) are much denser than others. For example, in WordNet (Miller, 1990; Fellbaum, 1998), a widely used, broad-coverage semantic network for English, it is not at all difficult to find links that cover an intuitively narrow distance (RABBIT EARS IS-A TELEVISION ANTENNA) or an intuitively wide one (PHYTOPLANKTON IS-A LIVING THING). The same kinds of examples can be found in the Collins COBUILD Dictionary (Sinclair, ed., 1987), which identifies superordinate terms for many words (e.g., SAFETY VALVE IS-A VALVE seems much narrower than KNITTING MACHINE IS-A MACHINE).

In the first part of this article, I describe an alternative way to evaluate semantic similarity in a taxonomy, based on the notion of information content. Like the edge-counting method, it is conceptually quite simple. However, it is not sensitive to the problem of varying link distances. In addition, by combining a taxonomic structure with empirical probability estimates, it provides a way of adapting a static knowledge structure to multiple contexts. Section 2 sets up the probabilistic framework and defines the measure of semantic similarity in information-theoretic terms, and Section 3 presents an evaluation of the similarity measure against human similarity judgments, using the simple edge-counting method as a baseline.

In the second part of the article, Sections 4 and 5, I describe two applications of semantic similarity to problems of ambiguity in natural language. The first concerns a particular case of syntactic ambiguity that involves both coordination and nominal compounds, each of which is a pernicious source of structural ambiguity in English. Consider the phrase *food handling and storage procedures*: does it represent a conjunction of *food handling* and *storage procedures*, or does it refer to the *handling and storage* of food? The second application concerns the resolution of word sense ambiguity — not for words in running text, which is a large open problem (though cf. Wilks & Stevenson, 1996), but for groups of related words as are often discovered by distributional analysis of text corpora or found in dictionaries and thesauri. Finally, Section 6 discusses related work.

## 2. Similarity and Information Content

Let $\mathcal{C}$ be the set of concepts in an IS-A taxonomy, permitting multiple inheritance. Intuitively, one key to the similarity of two concepts is the extent to which they share information, indicated in an IS-A taxonomy by a highly specific concept that subsumes them both. The edge-counting method captures this indirectly, since if the minimal path of IS-A links between two nodes is long, that means it is necessary to go high in the taxonomy, to more abstract concepts, in order to find a least upper bound. For example, in WordNet, NICKEL and DIME are both subsumed by COIN, whereas the most specific superclass that NICKEL and CREDIT CARD share is MEDIUM OF EXCHANGE (see Figure 1). In a feature-based setting (e.g., Tversky, 1977), this would be reflected by explicit shared features: nickels and dimes are both small, round, metallic, and so on. These features are captured implicitly by the taxonomy in categorizing NICKEL and DIME as subordinates of COIN.

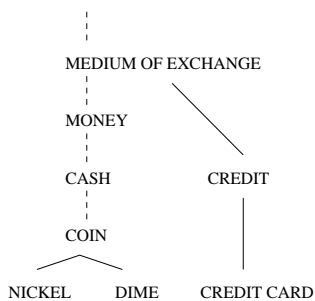

Figure 1: Fragment of the WordNet taxonomy. Solid lines represent IS-A links; dashed lines indicate that some intervening nodes were omitted to save space.

By associating probabilities with concepts in the taxonomy, it is possible to capture the same idea as edge-counting, but avoiding the unreliability of edge distances. Let the taxonomy be augmented with a function $p : \mathcal{C} \to [0, 1]$, such that for any $c \in \mathcal{C}$, $p(c)$ is the probability of encountering an instance of concept $c$. This implies that p is monotonically nondecreasing as one moves up the taxonomy: if $c_1$ IS-A $c_2$, then $p(c_1) \leq p(c_2)$. Moreover, if the taxonomy has a unique top node then its probability is 1.

Following the standard argumentation of information theory (Ross, 1976), the *information content* of a concept $c$ can be quantified as negative the log likelihood, $-\log p(c)$. Notice that quantifying information content in this way makes intuitive sense in this setting: as probability increases, informativeness decreases; so the more abstract a concept, the lower its information content. Moreover, if there is a unique top concept, its information content is 0.

This quantitative characterization of information provides a new way to measure semantic similarity. The more information two concepts share, the more similar they are, and the information shared by two concepts is indicated by the information content of the concepts that subsume them in the taxonomy. Formally, define

$$\mathrm{sim}(c_1, c_2) \quad = \quad \max_{c \in S(c_1, c_2)} \left[-\log p(c)\right], \tag{1}$$

where $S(c_1, c_2)$ is the set of concepts that subsume both $c_1$ and $c_2$. A class that achieves the maximum value in Equation 1 will be termed a *most informative subsumer*; most often there is a unique most informative subsumer, although this need not be true in the general case. Taking the maximum with respect to information content is analogous to taking the first intersection in semantic network marker-passing or the shortest path with respect to edge distance (cf. Quillian, 1968; Rada et al., 1989); a generalization from taking the maximum to taking a weighted average is introduced in Section 3.4.

Notice that although similarity is computed by considering all upper bounds for the two concepts, the information measure has the effect of identifying minimal upper bounds, since no class is less informative than its superordinates. For example, in Figure 1, COIN, CASH, etc. are all members of $S(\text{NICKEL}, \text{DIME})$, but the concept that is structurally the minimal upper bound, COIN, will also be the most informative. This can make a difference in cases of multiple inheritance: two distinct ancestor nodes may both be minimal upper bounds, as measured using distance in the graph, but those two nodes might have very different values for information content. Also notice that in IS-A taxonomies such as WordNet, where there are multiple sub-taxonomies but no unique top node, asserting zero similarity for concepts in separate sub-taxonomies (e.g., LIBERTY, AORTA) is equivalent to unifying the sub-taxonomies by creating a virtual topmost concept.

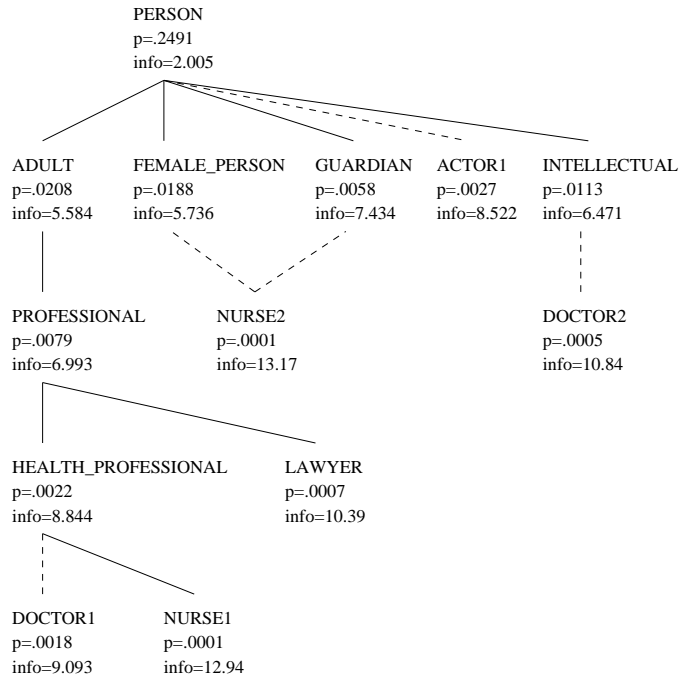

Figure 2: Another fragment of the WordNet taxonomy

In practice, one often needs to measure *word similarity*, rather than concept similarity. Using $s(w)$ to represent the set of concepts in the taxonomy that are senses of word $w$, define

$$\text{wsim}(w_1, w_2) \quad = \quad \max_{c1, c2} \left[\text{sim}(c_1, c_2)\right], \tag{2}$$

where $c_1$ ranges over $s(w_1)$ and $c_2$ ranges over $s(w_2)$. This is consistent with Rada et al.'s (1989) treatment of "disjunctive concepts" using edge-counting: they define the distance between two disjunctive sets of concepts as the minimum path length from any element of the first set to any element of the second. Here, the word similarity is judged by taking the maximal information content over all concepts of which both words could be an instance. To take an example, consider how the word similarity wsim(*doctor*, *nurse*) would be computed, using the taxonomic information in Figure 2. (Note that only noun senses are considered here.) By Equation 2, we must consider all pairs of concepts $\langle c_1, c_2 \rangle$, where $c_1 \in \{\text{DOCTOR1}, \text{DOCTOR2}\}$ and $c_2 \in \{\text{NURSE1}, \text{NURSE2}\}$, and for each such pair we must compute the semantic similarity $\text{sim}(c_1, c_2)$ according to Equation 1. Table 1 illustrates the computation.

| $c_1$ (description) | $c_2$ (description) | subsumer | sim($c_1$,$c_2$) |
|---|---|---|---|
| DOCTOR1 (medical) | NURSE1 (medical) | HEALTH_PROFESSIONAL | 8.844 |
| DOCTOR1 (medical) | NURSE2 (nanny) | PERSON | 2.005 |
| DOCTOR2 (Ph.D.) | NURSE1 (medical) | PERSON | 2.005 |
| DOCTOR2 (Ph.D.) | NURSE2 (nanny) | PERSON | 2.005 |

Table 1: Computation of similarity for *doctor* and *nurse*

As the table shows, when all the senses for *doctor* are considered against all the senses for *nurse*, the maximum value is 8.844, via HEALTH_PROFESSIONAL as a most informative subsumer; this is, therefore, the value of word similarity for *doctor* and *nurse*.[1]

## 3. Evaluation

This section describes a simple, direct method for evaluating semantic similarity, using human judgments as the basis for comparison.

### 3.1 Implementation

The work reported here used WordNet's taxonomy of concepts represented by nouns (and compound nominals) in English.[2] Frequencies of concepts in the taxonomy were estimated using noun frequencies from the Brown Corpus of American English (Francis & Kučera, 1982), a large (1,000,000 word) collection of text across genres ranging from news articles to science fiction. Each noun that occurred in the corpus was counted as an occurrence of each taxonomic class containing it.[3] For example, in Figure 1, an occurrence of the noun *dime* would be counted toward the frequency of DIME, COIN, CASH, and so forth. Formally,

$$\text{freq}(c) \;=\; \sum_{n \in \text{words}(c)} \text{count}(n), \qquad (3)$$

where words($c$) is the set of words subsumed by concept $c$. Concept probabilities were computed simply as relative frequency:

$$\hat{\text{p}}(c) \;=\; \frac{\text{freq}(c)}{N}, \qquad (4)$$

where $N$ was the total number of nouns observed (excluding those not subsumed by any WordNet class, of course). Naturally the frequency estimates in Equation 3 would be improved by taking into account the intended sense of each noun in the corpus — for example, an instance of *crane* can be a bird or a machine, but not both. Sense-tagged corpora are generally not available, however, and so the frequency estimates are done using this weaker but more generally applicable technique.

---

1. The taxonomy in Figure 2 is a fragment of WordNet version 1.6, showing real quantitative information computed using the method described below. The "nanny" sense of *nurse* (nursemaid, a woman who is the custodian of children) is primarily a British usage. The example omits two other senses of *doctor* in WordNet: a theologian in the Roman Catholic Church, and a game played by children. WordNet does not use node labels like DOCTOR1, but I have created such labels here for the sake of readability.

2. *Concept* as used here refers to what Miller et al. (1990) call a *synset*, essentially a node in the taxonomy. The experiment reported in this section used the noun taxonomy from WordNet version 1.4, which has approximately 50,000 nodes.

3. Plural nouns counted as instances of their singular forms.

It should be noted that the present method of associating probabilities with concepts in a taxonomy is not based on the notion of a single random variable ranging over all concepts — were that the case, the "credit" for each noun occurrence would be distributed over all concepts for the noun, and the counts normalized across the entire taxonomy to sum to 1. (That is the approach taken in Resnik, 1993a, also see Resnik, 1998b for discussion.) In assigning taxonomic probabilities for purposes of measuring semantic similarity, the present model associates a separate, binomially distributed random variable with each concept.[4] That is, from the perspective of any given concept $c$, an observed noun either is or is not an instance of that concept, with probabilities $p(c)$ and $1 - p(c)$, respectively. Unlike a model in which there is a single multinomial variable ranging over the entire set of concepts, this formulation assigns probability 1 to the top concept of the taxonomy, leading to the desirable consequence that its information content is zero.

## 3.2 Task

Although there is no standard way to evaluate computational measures of semantic similarity, one reasonable way to judge would seem to be agreement with human similarity ratings. This can be assessed by using a computational similarity measure to rate the similarity of a set of word pairs, and looking at how well its ratings correlate with human ratings of the same pairs.

An experiment by Miller and Charles (1991) provided appropriate human subject data for the task. In their study, 38 undergraduate subjects were given 30 pairs of nouns that were chosen to cover high, intermediate, and low levels of similarity (as determined using a previous study, Rubenstein & Goodenough, 1965), and those subjects were asked to rate "similarity of meaning" for each pair on a scale from 0 (no similarity) to 4 (perfect synonymy). The average rating for each pair thus represents a good estimate of how similar the two words are, according to human judgments.[5]

In order to get a baseline for comparison, I replicated Miller and Charles's experiment, giving ten subjects the same 30 noun pairs. The subjects were all computer science graduate students or postdoctoral researchers at the University of Pennsylvania, and the instructions were exactly the same as used by Miller and Charles, the main difference being that in this replication the subjects completed the questionnaire by electronic mail (though they were instructed to complete the whole task in a single uninterrupted sitting). Five subjects received the list of word pairs in a random order, and the other five received the list in the reverse order. The correlation between the Miller and Charles mean ratings and the mean ratings in my replication was .96, quite close to the .97 correlation that Miller and Charles obtained between their results and the ratings determined by the earlier study.

---

4. This is similar in spirit to the way probabilities are used in a Bayesian network.

5. An anonymous reviewer points out that human judgments on this task may be influenced by prototypicality, e.g., the pair *bird/robin* would likely yield higher ratings than *bird/crane*. Issues of this kind are briefly touched on in Section 6, but for the most part they are ignored here since prototypicality, like topical relatedness, is not captured in most IS-A taxonomies.

For each subject in my replication, I computed how well his or her ratings correlated with the Miller and Charles ratings. The average correlation over the 10 subjects was $r = 0.88$, with a standard deviation of $0.08$.[6] This value represents an upper bound on what one should expect from a computational attempt to perform the same task.

For purposes of evaluation, three computational similarity measures were used. The first is the similarity measurement using information content proposed in the previous section. The second is a variant on the edge-counting method, converting it from distance to similarity by subtracting the path length from the maximum possible path length:

$$\text{wsim}_{\text{edge}}(w_1, w_2) = (2 \times \text{MAX}) - \left[ \min_{c_1, c_2} \text{len}(c_1, c_2) \right] \tag{5}$$

where $c_1$ ranges over $s(w_1)$, $c_2$ ranges over $s(w_2)$, MAX is the maximum depth of the taxonomy, and $\text{len}(c_1, c_2)$ is the length of the shortest path from $c_1$ to $c_2$. (Recall that $s(w)$ denotes the set of concepts in the taxonomy that represent senses of word $w$.) If all senses of $w_1$ and $w_2$ are in separate sub-taxonomies of WordNet their similarity is taken to be zero. Note that because correlation is used as the evaluation metric, the conversion from a distance to a similarity can be viewed as an expository convenience, and does not affect the results: although the sign of the correlation coefficient changes from positive to negative, its magnitude turns out to be just the same regardless of whether or not the minimum path length is subtracted from $(2 \times \text{MAX})$.

The third point of comparison is a measure that simply uses the probability of a concept, rather than the information content, to define semantic similarity of concepts

$$\text{sim}_{\text{p}(c)}(c_1, c_2) = \max_{c \in S(c_1, c_2)} [1 - \text{p}(c)] \tag{6}$$

and the corresponding measure of word similarity:

$$\text{wsim}_{\text{p}(c)}(w_1, w_2) = \max_{c_1, c_2} \left[ \text{sim}_{\text{p}(c)}(c_1, c_2) \right], \tag{7}$$

where $c_1$ ranges over $s(w_1)$ and $c_2$ ranges over $s(w_2)$ in Equation 7. The probability-based similarity score is included in order to assess the extent to which similarity judgments might be sensitive to frequency *per se* rather than information content. Again, the difference between maximizing $1 - \text{p}(c)$ and minimizing $\text{p}(c)$ turns out not to affect the magnitude of the correlation. It simply ensures that the value can be interpreted as a similarity value, with high values indicating similar words.

### 3.3 Results

Table 2 summarizes the experimental results, giving the correlation between the similarity ratings and the mean ratings reported by Miller and Charles. Note that, owing to a noun missing from the WordNet 1.4 taxonomy, it was only possible to obtain computational similarity ratings for 28 of the 30 noun pairs; hence the proper point of comparison for human judgments is not the correlation over all 30 items ($r = .88$), but rather the correlation over the 28 included pairs ($r = .90$). The similarity ratings by item are given in Table 3.

---

6. Inter-subject correlation in the replication, estimated using leaving-one-out resampling (Weiss & Kulikowski, 1991), was $r = .90$, $\text{stdev} = 0.07$.

| Similarity method | Correlation |
|---|---|
| Human judgments (replication) | $r = .9015$ |
| Information content | $r = .7911$ |
| Probability | $r = .6671$ |
| Edge-counting | $r = .6645$ |

Table 2: Summary of experimental results.

| Word Pair | | Miller and Charles means | Replication means | wsim | wsim$_{edge}$ | wsim$_{p(c)}$ |
|---|---|---|---|---|---|---|
| car | automobile | 3.92 | 3.9 | 8.0411 | 30 | 0.9962 |
| gem | jewel | 3.84 | 3.5 | 14.9286 | 30 | 1.0000 |
| journey | voyage | 3.84 | 3.5 | 6.7537 | 29 | 0.9907 |
| boy | lad | 3.76 | 3.5 | 8.4240 | 29 | 0.9971 |
| coast | shore | 3.70 | 3.5 | 10.8076 | 29 | 0.9994 |
| asylum | madhouse | 3.61 | 3.6 | 15.6656 | 29 | 1.0000 |
| magician | wizard | 3.50 | 3.5 | 13.6656 | 30 | 0.9999 |
| midday | noon | 3.42 | 3.6 | 12.3925 | 30 | 0.9998 |
| furnace | stove | 3.11 | 2.6 | 1.7135 | 23 | 0.6951 |
| food | fruit | 3.08 | 2.1 | 5.0076 | 27 | 0.9689 |
| bird | cock | 3.05 | 2.2 | 9.3139 | 29 | 0.9984 |
| bird | crane | 2.97 | 2.1 | 9.3139 | 27 | 0.9984 |
| tool | implement | 2.95 | 3.4 | 6.0787 | 29 | 0.9852 |
| brother | monk | 2.82 | 2.4 | 2.9683 | 24 | 0.8722 |
| crane | implement | 1.68 | 0.3 | 2.9683 | 24 | 0.8722 |
| lad | brother | 1.66 | 1.2 | 2.9355 | 26 | 0.8693 |
| journey | car | 1.16 | 0.7 | 0.0000 | 0 | 0.0000 |
| monk | oracle | 1.10 | 0.8 | 2.9683 | 24 | 0.8722 |
| food | rooster | 0.89 | 1.1 | 1.0105 | 18 | 0.5036 |
| coast | hill | 0.87 | 0.7 | 6.2344 | 26 | 0.9867 |
| forest | graveyard | 0.84 | 0.6 | 0.0000 | 0 | 0.0000 |
| monk | slave | 0.55 | 0.7 | 2.9683 | 27 | 0.8722 |
| coast | forest | 0.42 | 0.6 | 0.0000 | 0 | 0.0000 |
| lad | wizard | 0.42 | 0.7 | 2.9683 | 26 | 0.8722 |
| chord | smile | 0.13 | 0.1 | 2.3544 | 20 | 0.8044 |
| glass | magician | 0.11 | 0.1 | 1.0105 | 22 | 0.5036 |
| noon | string | 0.08 | 0.0 | 0.0000 | 0 | 0.0000 |
| rooster | voyage | 0.08 | 0.0 | 0.0000 | 0 | 0.0000 |

Table 3: Semantic similarity by item.

| n1 | n2 | wsim($n_1,n_2$) | subsumer |
|--------|---------|-------:|-----------|
| tobacco | alcohol | 7.63 | DRUG |
| tobacco | sugar | 3.56 | SUBSTANCE |
| tobacco | horse | 8.26 | NARCOTIC |

Table 4: Similarity with *tobacco* computed by maximizing information content

### 3.4 Discussion

The experimental results in the previous section suggest that measuring semantic similarity using information content provides results that are better than the traditional method of simply counting the number of intervening IS-A links.

The measure is not without its problems, however. Like simple edge-counting, the measure sometimes produces spuriously high similarity measures for words on the basis of inappropriate word senses. For example, Table 4 shows the word similarity for several words with *tobacco*. *Tobacco* and *alcohol* are similar, both being drugs, and *tobacco* and *sugar* are less similar, though not entirely dissimilar, since both can be classified as substances. The problem arises, however, in the similarity rating for *tobacco* with *horse*: the word *horse* can be used as a slang term for *heroin*, and as a result information-based similarity is maximized, and path length minimized, when the two words are both categorized as narcotics. This is contrary to intuition.

Cases like this are probably relatively rare. However, the example illustrates a more general concern: in measuring similarity between words, it is really the relationship among word *senses* that matters, and a similarity measure should be able to take this into account.

In the absence of a reliable algorithm for choosing the appropriate word senses, the most straightforward way to do so in the information-based setting is to consider *all* concepts to which both nouns belong rather than taking just the single maximally informative class. This suggests defining a measure of *weighted word similarity* as follows:

$$\text{wsim}_\alpha(w_1, w_2) \quad = \quad \sum_i \alpha(c_i)[-\log \text{p}(c_i)], \qquad (8)$$

where $\{c_i\}$ is the set of concepts dominating both $w_1$ and $w_2$ in any sense of either word, and $\alpha$ is a weighting function over concepts such that $\sum_i \alpha(c_i) = 1$. This measure of similarity takes more information into account than the previous one: rather than relying on the single concept with *maximum* information content, it allows *each* class representing shared properties to contribute information content according to the value of $\alpha(c_i)$. Intuitively, these $\alpha$ values measure relevance. For example, in computing $\text{wsim}_\alpha$(*tobacco,horse*), the $c_i$ would range over all concepts of which *tobacco* and *horse* are both instances, including NARCOTIC, DRUG, ARTIFACT, LIFE_FORM, etc. In an everyday context one might expect low values for $\alpha$(NARCOTIC) and $\alpha$(DRUG), but in the context of, say, a newspaper article about drug dealers, the weights of these concepts might be quite high. Although it is not possible to include weighted word similarity in the comparison of Section 3, since the noun pairs are judged without context, Section 4 provides further discussion and a weighting function $\alpha$ designed for a particular natural language processing task.

## 4. Using Taxonomic Similarity in Resolving Syntactic Ambiguity

Having considered a direct evaluation of the information-based semantic similarity measure, I now turn to the application of the measure in resolving syntactic ambiguity.

### 4.1 Clues for Resolving Coordination Ambiguity

Syntactic ambiguity is a pervasive problem in natural language. As Church and Patil (1982) point out, the class of "every way ambiguous" syntactic constructions — those for which the number of analyses is the number of binary trees over the terminal elements — includes such frequent constructions as prepositional phrases, coordination, and nominal compounds. In the last several years, researchers in natural language have made a great deal of progress in using quantitative information from text corpora to provide the needed constraints. Progress on broad-coverage prepositional phrase attachment ambiguity has been particularly notable, now that the dominant approach has shifted from structural strategies to quantitative analysis of lexical relationships (Whittemore, Ferrara, & Brunner, 1990; Hindle & Rooth, 1993; Brill & Resnik, 1994; Ratnaparkhi & Roukos, 1994; Li & Abe, 1995; Collins & Brooks, 1995; Merlo, Crocker, & Berthouzoz, 1997). Noun compounds have received comparatively less attention (Kobayasi, Takunaga, & Tanaka, 1994; Lauer, 1994, 1995), as has the problem of coordination ambiguity (Agarwal & Boggess, 1992; Kurohashi & Nagao, 1992).

In this section, I investigate the role of semantic similarity in resolving coordination ambiguities involving nominal compounds. I began with noun phrase coordinations of the form *n1 and n2 n3*, which admit two structural analyses, one in which *n1* and *n2* are the two noun phrase heads being conjoined (1a) and one in which the conjoined heads are *n1* and *n3* (1b).

(1)   a. a (bank and warehouse) guard
      b. a (policeman) and (park guard)

Identifying which two head nouns are conjoined is necessary in order to arrive at a correct interpretation of the phrase's content. For example, analyzing (1b) according to the structure of (1a) could lead a machine translation system to produce a noun phrase describing somebody who guards both policemen and parks. Analyzing (1a) according to the structure of (1b) could lead an information retrieval system to miss this phrase when looking for queries involving the term *bank guard.*

Kurohashi and Nagao (1992) point out that similarity of form and similarity of meaning are important cues to conjoinability. In English, similarity of form is to a great extent captured by agreement in number (singular vs. plural):

(2)   a. several *business* and *university* groups
      b. several *businesses* and university *groups*

Similarity of form between candidate conjoined heads can thus be thought of as a Boolean variable: number agreement is either satisfied by the candidate heads or it is not.

Similarity of meaning of the conjoined heads also appears to play an important role:

(3)   a. a *television* and *radio* personality

     b. a *psychologist* and sex *researcher*

Clearly *television* and *radio* are more similar than *television* and *personality*; corresponding for *psychologist* and *researcher*. This similarity of meaning is captured well by semantic similarity in a taxonomy, and thus a second variable to consider when evaluating a coordination structure is semantic similarity as measured by overlap in information content between the two head nouns.

In addition, for the constructions considered here, the appropriateness of noun-noun modification is relevant:

(4)    a. *mail* and *securities* fraud
       b. *corn* and peanut *butter*

One reason we prefer to conjoin *mail* and *securities* is that *mail fraud* is a salient compound nominal phrase. On the other hand, *corn butter* is not a familiar concept; compare to the change in perceived structure if the phrase were *corn and peanut crops*. In order to measure the appropriateness of noun-noun modification, I use a quantitative measure of selectional fit called *selectional association* (Resnik, 1996), which takes into account both lexical co-occurrence frequencies and semantic class membership in the WordNet taxonomy. Briefly, the selectional association of a word $w$ with a WordNet class $c$ is given by

$$A(w, c) \;=\; \frac{\mathrm{p}(c|w) \log \frac{\mathrm{p}(c|w)}{\mathrm{p}(c)}}{\mathrm{D}(\mathrm{p}(C|w) \parallel \mathrm{p}(C))} \tag{9}$$

where $\mathrm{D}(p_1 \parallel p_2)$ is the Kullback-Leibler distance (relative entropy) between probability distributions $p_1$ and $p_2$. Intuitively, $A(w, c)$ is measuring the extent to which class $c$ is predicted by word $w$; for example, $A(wool, \text{CLOTHING})$ would have a higher value than, say, $A(wool, \text{PERSON})$. The selectional association $A(w_1, w_2)$ of two words is defined as the maximum of $A(w_1, c)$ taken over all classes $c$ to which $w_2$ belongs. For example, $A(wool, glove)$ would most likely be equal to $A(wool, \text{CLOTHING})$, as compared to, say, $A(wool, \text{SPORTS\_EQUIPMENT})$ — the latter value corresponding to the sense of *glove* as something used in baseball or in boxing. (See Li & Abe, 1995, for an approach in which selectional relationships are modeled using conditional probability.) A simple way to treat selectional association as a variable in resolving coordination ambiguities is to prefer analyses that include noun-noun modifications with very strong affinities (e.g., *bank* as a modifier of *guard*) and to disprefer very weak noun-noun relationships (e.g., *corn* as a modifier of *butter*). Thresholds defining "strong" and "weak" are parameters of the algorithm, defined below.

## 4.2 Resolving Coordination Ambiguity: First Experiment

I investigated the roles of these sources of evidence by conducting a straightforward disambiguation experiment using naturally occurring linguistic data. Two sets of 100 noun phrases of the form [NP *n1 and n2 n3*] were extracted from the parsed *Wall Street Journal* (WSJ) corpus, as found in the Penn Treebank (Marcus, Santorini, & Marcinkiewicz, 1993). These were disambiguated by hand, with one set used for development and the other for testing.[7] A set of simple transformations were applied to all WSJ data, including the mapping of all proper names to the token *someone*, the expansion of month abbreviations, and the reduction of all nouns to their root forms.

| Source of evidence | Conjoined | Condition |
|---|---|---|
| Number agreement | n1 and n2 | number(n1) = number(n2) AND number(n1) ≠ number(n3) |
| | n1 and n3 | number(n1) = number(n3) AND number(n1) ≠ number(n2) |
| | undecided | otherwise |
| Semantic similarity | n1 and n2 | wsim(n1,n2) > wsim(n1,n3) |
| | n1 and n3 | wsim(n1,n3) > wsim(n1,n2) |
| | undecided | otherwise |
| Noun-noun modification | n1 and n2 | A(n1,n3) > $\tau$ OR A(n3,n1) > $\tau$ |
| | n1 and n3 | A(n1,n3) < $\sigma$ OR A(n3,n1) < $\sigma$ |
| | undecided | otherwise |

Table 5: Rules for number agreement, semantic similarity, and noun-noun modification in resolving syntactic ambiguity of noun phrases *n1 and n2 n3*

Number agreement was determined using a simple analysis of suffixes in combination with WordNet's lists of root nouns and irregular plurals.[8] Semantic similarity was determined using the information-based measure of Equation (2) — the noun class probabilities of Equation (1) were estimated using a sample of approximately 800,000 noun occurrences in Associated Press newswire stories.[9] For the purpose of determining semantic similarity, nouns not in WordNet were treated as instances of the class ⟨thing⟩. Appropriateness of noun-noun modification was determined by computing selectional association (Equation 9), using co-occurrence frequencies taken from a sample of approximately 15,000 noun-noun compounds extracted from the WSJ corpus. (This sample did not include the test data.) Both selection of the modifier for the head and selection of the head for the modifier were considered by the disambiguation algorithm. Table 5 provides details of the decision rule for each source of evidence when used independently.[10]

In addition, I investigated several methods for combining the three sources of information. These included: (a) a simple form of "backing off" (specifically, given the number agreement, noun-noun modification, and semantic similarity strategies in that order, use the choice given by the first strategy that isn't undecided); (b) taking a vote among the three strategies and choosing the majority; (c) classifying using the results of a linear re-gression; and (d) constructing a decision tree classifier. The latter two methods are forms of supervised learning; in this experiment the development set was used as the training data.[11]

---

7. Hand disambiguation was necessary because the Penn Treebank does not encode NP-internal structure. These phrases were disambiguated using the full sentence in which they occurred, plus the previous and following sentence, as context.

8. The experiments in this section used WordNet version 1.2.

9. I am grateful to Donald Hindle for making these data available.

10. Thresholds $\tau = 2.0$ and $\sigma = 0.0$ were fixed manually based on experience with the development set before evaluating the test data.

| Strategy | Coverage (%) | Accuracy (%) |
|---|---|---|
| Default | 100.0 | 66.0 |
| Number agreement | 53.0 | 90.6 |
| Noun–noun modification | 75.0 | 69.3 |
| Semantic similarity | 66.0 | 71.2 |
| Backing off | 95.0 | 81.1 |
| Voting | 89.0 | 78.7 |
| Number agreement + DEFAULT | 100.0 | 82.0 |
| Noun–noun modification + DEFAULT | 100.0 | 65.0 |
| Semantic similarity + DEFAULT | 100.0 | 72.0 |
| Backing off + DEFAULT | 100.0 | 81.0 |
| Voting + DEFAULT | 100.0 | 76.0 |
| Regression | 100.0 | 79.0 |
| ID3 Tree | 100.0 | 80.0 |

Table 6: Syntactic disambiguation for items of the form *n1 and n2 n3*

The results are shown in Table 6. The development set contained a bias in favor of conjoining *n1* and *n2*; therefore a "default" strategy, always choosing that bracketing, was used as a baseline for comparison. The default was also used for resolving undecided cases in order to make comparisons of individual strategies at 100% coverage. For example, "Number agreement + DEFAULT" shows the figures obtained when number agreement is used to make the choice and the default is selected if that choice is undecided.

Not surprisingly, the individual strategies perform reasonably well on the instances they can classify, but coverage is poor; the strategy based on similarity of form is the most highly accurate, but arrives at an answer only half the time. However, the heavy a priori bias makes up the difference — to such an extent that even though the other forms of evidence have some value, no combination beats the number-agreement-plus-default combination. On the positive side, this shows that the ambiguity can be resolved reasonably well using a very simple algorithm: viewed in terms of how many errors are made, number agreement makes it possible to cut the baseline 34% error rate nearly in half to 18% incorrect analyses (a 44% reduction). On the negative side, the results fail to make a strong case that semantic similarity can add something useful.

Before taking up this issue, let us assess the contributions of the individual strategies to the results when evidence is combined, by further analyzing the behavior of the unsupervised evidence combination strategies. When combining evidence by voting, a choice was made in 89 cases. The number agreement strategy agreed with the majority vote in 57 cases, of which 43 (75.4%) were correct; the noun-noun modification strategy agreed with the majority in 73 cases, of which 50 (68.5%) were correct; and the semantic similarity strategy agreed with the majority in 58 cases, of which 43 (74.1%) were correct. In the "backing off" form of evidence combination, number agreement makes a choice for 53 cases and is correct for 48 (90.6%); then, of those remaining undecided, noun-noun modification makes a choice for 35 cases and is correct for 24 (68.6%); then, of those still undecided, semantic similarity makes a choice for 7 cases of which 5 are correct (71.4%); and the remaining 5 cases are undecided.

---

11. What I am calling "backing off" is related in spirit to Katz's well known smoothing technique (Katz, 1987), but the "backing off" strategy used here is not quantitative. I retain the double quotes in order to highlight the distinction.

This analysis and the above-baseline performance of the semantic-similarity-plus-default strategy show that semantic similarity does contain information about the correct answer: it agrees with the majority vote a substantial portion of the time, and it selects correct answers more often than one would expect by default for the cases it receives through "backing off." However, because the default is correct two thirds of the time, and because the number agreement strategy is correct nine out of ten times for the cases it can decide, the potential contribution of semantic similarity remains suggestive rather than conclusive. In a second experiment, therefore, I investigated a more difficult formulation of the problem in order to obtain a better assessment.

## 4.3 Resolving Coordination Ambiguity: Second Experiment

In the second experiment using the same data sources, I investigated a more complex set of coordinations, looking at noun phrases of the form *n0 n1 and n2 n3*. The syntactic analyses of such phrases are characterized by the same top-level binary choice as the data in the previous experiment, either conjoining heads *n1* and *n2* as in (5) or conjoining *n1* and *n3* as in (6).[12]

(5)    a. freshman ((business and marketing) major)
      b. (food (handling and storage)) procedures
      c. ((mail fraud) and bribery) charges

(6)    a. Clorets (gum and (breath mints))
      b. (baby food) and (puppy chow)

For this experiment, one set of 89 items was extracted from the Penn Treebank WSJ data for development, and another set of 89 items was set aside for testing. The development set showed significantly less bias than the data in the previous experiment, with 53.9% of items conjoining *n1* and *n2*.

The disambiguation strategies in this experiment were a more refined version of those used in the previous experiment, as illustrated in Table 7. Number agreement was used just as before. However, rather than employing semantic similarity and noun-noun modification as independent strategies — something not clearly warranted given the lackluster performance of the modification strategy — the two were combined in a measure of weighted semantic similarity as defined in Equation (8). Selectional association was used as the basis for $\alpha$. In particular, $\alpha_{1,2}(c)$ was the greater of A(n0,c) and A(n3,c), capturing the fact that when *n1* and *n2* are conjoined, the combined phrase potentially stands in a head-modifier relationship with *n0* and a modifier-head relationship with *n3*. Correspondingly, $\alpha_{1,3}(c)$ was the greater of A(n0,c) and A(n2,c), capturing the fact that the coordination of *n1* and *n3* takes place in the context of *n2* modifying *n3* and of *n1* (or a coordinated phrase containing it) being modified by *n0*.

---

12. The full 5-way classification problem for the structures in (5) and (6) was not investigated.

| Source of evidence | Conjoined | Condition |
|---|---|---|
| Number agreement | n1 and n2 | number(n1) = number(n2) AND number(n1) ≠ number(n3) |
| | n1 and n3 | number(n1) = number(n3) AND number(n1) ≠ number(n2) |
| | undecided | otherwise |
| Weighted semantic similarity | n1 and n2 | $\text{wsim}_{\alpha_{1,2}}(\text{n1,n2}) > \text{wsim}_{\alpha_{1,3}}(\text{n1,n3})$ |
| | n1 and n3 | $\text{wsim}_{\alpha_{1,3}}(\text{n1,n3}) > \text{wsim}_{\alpha_{1,2}}(\text{n1,n2})$ |
| | undecided | otherwise |

Table 7: Rules for number agreement and weighted semantic similarity in resolving syntactic ambiguity of noun phrases *n0 n1 and n2 n3*

| STRATEGY | COVERAGE (%) | ACCURACY (%) |
|---|---|---|
| Default | 100.0 | 44.9 |
| Number agreement | 40.4 | 80.6 |
| Weighted semantic similarity | 69.7 | 77.4 |
| Backing off | 85.4 | 81.6 |

Table 8: Syntactic disambiguation for items of the form *n0 n1 and n2 n3*

For example, consider an instance of the ambiguous phrase:

(7)    telecommunications products and services units.

It so happens that a high-information-content connection exists between *product* in its sense as "a quantity obtained by multiplication" and *unit* in its sense as "a single undivided whole." As a result, although neither of these senses is relevant for this example, nouns *n1* and *n3* would be assigned a high value for (unweighted) semantic similarity and be chosen incorrectly as the conjoined heads for this example. However, the unweighted similarity computation misses an important piece of context: in any syntactic analysis conjoining *product* and *unit* (cf. examples 6a and 6b), the word *telecommunications* is necessarily a modifier of the concept identified by *products*. But the selectional association between *telecommunications* and *products* in its "multiplication" sense is weak or nonexistent. Weighting by selection association, therefore, provides a way to reduce the impact of the spurious senses on the similarity computation.

In order to combine sources of evidence, I used "backing off" (from number agreement to weighted semantic similarity) to combine the two individual strategies. As a baseline, results were evaluated against a simple default strategy of always choosing the group that was more common in the development set. The results are shown in Table 8.

In this case, the default strategy defined using the development set was misleading, yielding worse than chance accuracy. For this reason, strategy-plus-default figures are not reported. However, even if default choices were made using the bias found in the test set, accuracy would be only 55.1%. In contrast to the equivocal results in the first experiment, this experiment demonstrates a clear contribution of semantic similarity: by employing semantic similarity in those cases where the more accurate number-agreement strategy cannot apply, it is possible to obtain equivalent or even somewhat better accuracy than number agreement alone while at the same time more than doubling the coverage.

Comparison with previous algorithms is unfortunately not possible, since researchers on coordination ambiguity have not established a common data set for evaluation or even a common characterization of the problem, in contrast to the now-standard *(v, n1, prep, n2)* contexts used in work on propositional phrase attachment. With that crucial caveat, it is nonetheless interesting to note that the results obtained here are broadly consistent with Kurohashi and Nagao (1992), who report accuracy results in the range of 80-83% at 100% coverage when analyzing a broad range of conjunctive structures in Japanese using a combination of string matching, syntactic similarity, and thesaurus-based similarity, and with Agarwal and Boggess (1992), who use syntactic types and structure, along with partly domain-dependent semantic labels, to obtain accuracies in a similar range for identifying conjuncts in English.

## 5. Using Taxonomic Similarity in Word Sense Selection

This section considers the application of the semantic similarity measure in resolving another form of ambiguity: selecting the appropriate sense of a noun when it appears in the context of other nouns that are related in meaning.

### 5.1 Associating Word Senses with Noun Groupings

Knowledge about groups of related words plays a role in many natural language applications. As examples, query expansion using related words is a well studied technique in information retrieval (e.g., Harman, 1992; Grefenstette, 1992), clusters of similar words can play a role in smoothing stochastic language models for speech recognition (Brown, Della Pietra, deSouza, Lai, & Mercer, 1992), classes of verbs that share semantic structure form the basis for an approach to interlingual machine translation (Dorr, 1997), and clusterings of related words can be used in characterizing subgroupings of retrieved documents in large-scale Web searches (e.g., Digital Equipment Corporation, 1998). There is a wide body of research on the use of distributional methods for measuring word similarity in order to obtain groups of related words (e.g., Bensch & Savitch, 1992; Brill, 1991; Brown et al., 1992; Grefenstette, 1992, 1994; McKeown & Hatzivassiloglou, 1993; Pereira, Tishby, & Lee, 1993; Schütze, 1993), and thesauri such as WordNet are another source of word relationships (e.g., Voorhees, 1994).

Distributional techniques can sometimes do a good job of identifying groups of related words (see Resnik, 1998b, for an overview and critical discussion), but for some tasks the relevant relationships are not among words, but among word *senses*. For example, Brown et al. (1992) illustrate the notion of a distributionally derived, "semantically sticky" cluster using an automatically derived word group containing *attorney, counsel, trial, court,* and *judge.* Although the semantic coherence of this cluster "pops out" for a human reader, a naive computational system has no defense against word sense ambiguity: using this cluster for query expansion could result in retrieving documents involving advice (one sense of *counsel*) and royalty (as one sense of *court*).[13]

Resnik (1998a) introduces an algorithm that uses taxonomically-defined semantic similarity in order to derive grouping relationships among word senses from grouping relationships among words. Formally, the problem can be stated as follows. Consider a set of words $W = \{w_1, \ldots, w_n\}$, with each word $w_i$ having an associated set $S_i = \{s_{i,1}, \ldots, s_{i,m}\}$ of possible senses. Assume that there exists some set $W' \subseteq \bigcup S_i$, representing the set of word *senses* that an ideal human judge would conclude belong to the group of senses corresponding to the word grouping $W$. (It follows that $W'$ must contain at least one representative from each $S_i$.) The goal is then to define a membership function $\varphi$ that takes $s_{i,j}$, $w_i$, and $W$ as its arguments and computes a value in $[0, 1]$, representing the confidence with which one can state that sense $s_{i,j}$ belongs in sense grouping $W'$. Note that, in principle, nothing precludes the possibility that multiple senses of a word are included in $W'$.

For example, consider again the group

> attorney, counsel, trial, court, judge.

Restricting attention to noun senses in WordNet, every word but *attorney* is polysemous. Treating this word group as $W$, a good algorithm for computing $\varphi$ should assign a value of 1 to the unique sense of *attorney*, and it should assign a high value to the sense of *counsel* as

> a lawyer who pleads cases in court.

Similarly, it should assign high values to the senses of *trial* as

> legal proceedings consisting of the judicial examination of issues by a competent tribunal

> the determination of a person's innocence or guilt by due process of law.

It should also assign high values to the senses of *court* as

> an assembly to conduct judicial business

> a room in which a law court sits.

And it should assign a high value to the sense of *judge* as

> a public official authorized to decide questions brought before a court of justice.

It should assign low values of $\varphi$ to the various word senses of words in this cluster that are associated with the group to a lesser extent or not at all. These would include the sense of *counsel* as

> direction or advice as to a decision or course of action;

similarly, a low value of $\varphi$ should be assigned to other senses of *court* such as

---

13. See Krovetz and Kroft, 1992 and Voorhees, 1993 for experimentation and discussion of the effects of word sense ambiguity in information retrieval.

**Algorithm** (Resnik, 1998a). Given $W = \{w_1, \ldots, w_n\}$, a set of nouns:

```
for i and j = 1 to n, with i < j
{
    v_{i,j} = wsim(w_i, w_j)
    c_{i,j} = the most informative subsumer for w_i and w_j

    for k = 1 to num_senses(w_i)
        if c_{i,j} is an ancestor of sense_{i,k}
            increment support[i, k] by v_{i,j}

    for k' = 1 to num_senses(w_j)
        if c_{i,j} is an ancestor of sense_{j,k'}
            increment support[j, k'] by v_{i,j}

    increment normalization[i] by v_{i,j}
    increment normalization[j] by v_{i,j}
}

for i = 1 to n
    for k = 1 to num_senses(w_i)
    {
        if (normalization[i] > 0.0)
            φ_{i,k} = support[i, k] / normalization[i]
        else
            φ_{i,k} = 1 / num_senses(w_i)
    }
```

Figure 3: Disambiguation algorithm for noun groupings

a yard wholly or partly surrounded by walls or buildings.

The disambiguation algorithm for noun groups is given in Figure 3. Intuitively, when two polysemous words are similar, their most informative subsumer provides information about which sense of each word is the relevant one. This observation is similar in spirit to other approaches to word sense disambiguation based on maximizing relatedness of meaning (e.g., Lesk, 1986; Sussna, 1993). The key idea behind the algorithm is to consider the nouns in a word group pairwise. For each pair the algorithm goes through all possible combinations of the words' senses, and assigns "credit" to senses on the basis of shared information content, as measured using the information content of the most informative subsumer.[14]

As an example, WordNet lists *doctor* as meaning either a medical doctor or someone holding a Ph.D., and lists *nurse* as meaning either a health professional or a nanny, but when the two words are considered together, the medical sense of each word is obvious to the human reader. This effect finds its parallel in the operation of the algorithm. Given a taxonomy like that of Figure 2, consider a case in which the set $W$ of words contains $w_1 = doctor$, $w_2 = nurse$, and $w_3 = actor$. In the first pairwise comparison, for *doctor* and *nurse*, the most informative subsumer is $c_{1,2} =$ HEALTH PROFESSIONAL, which has information content $v_{1,2} = 8.844$. Therefore the support for DOCTOR1 and NURSE1 is incremented by 8.844. Neither DOCTOR2 nor NURSE2 receives any increment in support based on this comparison, since neither has HEALTH PROFESSIONAL as an ancestor. In the second pairwise comparison, the most informative subsumer for *doctor* and *actor* is $c_{1,3} =$ PERSON, with information content $v_{1,3} = 2.005$, and so there is an increment by that amount to the support for DOCTOR1, DOCTOR2, and ACTOR1, all of which have PERSON as an ancestor. Similarly, in the third pairwise comparison, the most informative subsumer for *nurse* and *actor* is also PERSON, so NURSE1, NURSE2, and ACTOR1 all have their support incremented by 2.005. In the end, therefore, DOCTOR1 has received support $8.884 + 2.005$ out of a possible $8.884 + 2.005$ for all the pairwise comparisons in which it participated, so for that word sense $\varphi = 1$. In contrast, DOCTOR2 received support in the amount of 2.005 out of a possible $8.884 + 2.005$ for the comparisons in which it was involved, so the value of $\varphi$ for DOCTOR2 is $\frac{2.005}{8.884+2.005} = 0.185$.

---

14. In Figure 3, the square bracket notation highlights the fact that `support` is a matrix and `normalization` is an array. Conceptually `v` and `c` are (triangular) matrices also; however, I use subscripts rather than square brackets because at implementation time there is no need to implement them as such since the values $v_{i,j}$ and $c_{i,j}$ are used and discarded on each pass through the double loop.

Resnik (1998a) illustrates the algorithm of Figure 3 using word groupings from a variety of sources, including several of the sources on distributional clustering cited above, and evaluates the algorithm more rigorously on the task of associating WordNet senses with nouns in Roget's thesaurus, based on their thesaurus category membership. On average, the algorithm achieved approximately 89% of the performance of human annotators performing the same task.[15] In the remainder of this section I describe a new application of the algorithm, and evaluate its performance.

## 5.2 Linking to WordNet using a Bilingual Dictionary

Multilingual resources for natural language processing can be difficult to obtain, although some promising efforts are underway in projects like EuroWordNet (Vossen, 1998). For many languages, however, such large-scale resources are unlikely to be available in the near future, and individual research efforts will have to continue to build from scratch or to adapt existing resources such as bilingual dictionaries (e.g., Klavans & Tzoukermann, 1995). In this section I describe an application of the algorithm of Figure 3 to the English definitions in the CETA Chinese-English dictionary (CETA, 1982). The ultimate task, being undertaken in the context of a Chinese-English machine translation project, will be to associate Chinese vocabulary items with nodes in WordNet, much in the same way that vocabulary in Spanish, Dutch, and Italian are associated with interlingual taxonomy nodes derived from the American WordNet, in the EuroWordNet project; the task is also similar to attempts to relate dictionaries and thesauri monolingually (e.g., see Section 5.3 and Ji, Gong, & Huang, 1998). The present study investigates the extent to which semantic similarity might be useful in partially automating the process.

---

15. The task was performed independently by two human judges. Treating Judge 1 as the benchmark the accuracies achieved by Judge 2, the algorithm, and random selection were respectively 65.7%, 58.6%, and 34.8%; treating Judge 2 as the benchmark the accuracies achieved by Judge 1, the algorithm, and random selection were respectively 68.6%, 60.5%, and 33.3%. As the relatively low accuracies for human judges demonstrate, disambiguation using WordNet's fine-grained senses is quite a bit more difficult than disambiguation to the level of homographs (Hearst, 1991; Cowie, Guthrie, & Guthrie, 1992). Resnik and Yarowsky (1997, 1999) discuss the implications of WordNet's fine-grainedness for evaluation of word sense disambiguation, and consider alternative evaluation methods.

For example, consider the following dictionary entries:

(a) 阿伯: 1. ⟨lit⟩ brother-in-law (husband's elder brother) 2. ⟨reg⟩ father 3. ⟨reg⟩ uncle (father's elder brother) 4. uncle (form of address to an older man)

(b) 唱旦的: actress, player of female roles.

In order to associate Chinese terms such as these with the WordNet noun taxonomy, it is important to avoid associations with inappropriate senses — for example, the word in entry (a), 阿伯, should clearly not be associated with *father* in its WordNet senses as Church Father, priest, God-the-Father, or founding father.[16]

Although one traditional approach to using dictionary entries has been to compute word overlap with respect to dictionary definitions (e.g., Lesk, 1986), the English glosses in the CETA dictionary are generally too short to take advantage of word overlap in this fashion. However, many of the definitions do have a useful property: they possess multiple sub-definitions that are similar in meaning, as in the cases illustrated above. Although one cannot always assume that this is so, e.g.,

(c) 字盘: 1. case (i.e., upper case or lower case) 2. dial (of a watch, etc.),

inspection of the dictionary confirms that when multiple definitions are present they tend more toward polysemy than homonymy.

Based on this observation, I conducted an experiment to assess the extent to which the word sense disambiguation algorithm of Figure 3 can be used to identify relevant noun senses in WordNet for Chinese words in the CETA dictionary, using the English definitions as the source of similar nouns to disambiguate. Nouns heading definitional noun phrases were extracted automatically via simple heuristic methods, for a randomly-selected sample of 100 dictionary entries containing multiple definitions to be used as a test set. For example, the noun groups associated with the definitions above would be

(a') uncle, brother-in-law, father

(b') actress, player.

WordNet's noun database was used to automatically identify compound nominals where possible. So, for example, a word defined as "record player" would have the compound *record_player* rather than *player* as its head noun because *record_player* is a compound noun known to WordNet.[17]

It should be noted that no attempt was made to exclude dictionary entries like (c) when creating the test set. Since in general there is no way to automatically identify alternative definitions distinguished by synonymy from those distinguished by homonymy, such entries must be faced by any disambiguation algorithm for this task.

Two independent judges were recruited for assistance in annotating the test set, one a native Chinese speaker, and the second a Chinese language expert for the United States government. These judges independently annotated the 100 test items. For each item,

---

16. Annotations within the dictionary entries such as <lit> (literary), <reg> (regional), and the like are ignored by the algorithm described in this section.

17. WordNet version 1.5 was used for this experiment.

For each WordNet definition, you will see 6 boxes: 1, 2, 3, 4, 5, and IS-A. For each definition:

- if you think the Chinese word *can have that meaning*, select the number corresponding to your confidence in that choice, where 1 is **lowest** confidence and 5 is **highest** confidence.

- If the Chinese word cannot have that meaning, but can have a *more specific* meaning, select IS-A. For example, if the Chinese word means "truck" and the WordNet definition is "automotive vehicle: self-propelled wheeled vehicle", you would select this option. (That is, it makes sense to say that this Chinese word describes a concept that IS A KIND OF AUTOMOTIVE VEHICLE.) **Then** pick 1, 2, 3, 4, or 5 as your confidence in this decision, again with 1 as **lowest** confidence and 5 as **highest** confidence.

- If neither of the above cases apply for this WordNet definition, don't check off anything for this definition.

Figure 4: Instructions for human judges selecting senses associated with Chinese words

the judge was given the Chinese word, its full CETA dictionary definition (as in examples a–c), and a list of all the WordNet sense descriptions associated with any sense of any head noun in the associated noun group. For example, the list corresponding to the following dictionary definition

(d) 急讯: urgent message, urgent dispatch

would contain the following WordNet sense descriptions, as generated via the head nouns *message* and *dispatch*:

- message, content, subject_matter, substance: what a communication that is about something is about

- dispatch, expedition, expeditiousness, fastness: subconcept of celerity, quickness, rapidity

- dispatch, despatch, communique: an official report (usually sent in haste)

- message: a communication (usually brief) that is written or spoken or signaled; "he sent a three-word message"

- dispatch, despatch, shipment: the act of sending off something

- dispatch, despatch: the murder or execution of someone

For each item, the judge was first asked whether he knew that Chinese word in that meaning; if the response was negative, he was instructed to proceed to the next item. For items with known words, the judges were instructed as in Figure 4.

Although the use of the IS-A selection was not used in the analysis of the results, it was important to include it because it provided the judges with a way to indicate where a Chinese word could best be classified in the WordNet noun taxonomy, without having to assert translational equivalence between the Chinese concept and a close WordNet (English) concept. So, for example, a judge could classify the word 春节 (the spring festival, lunar new year, Chinese new year) as belonging under the WordNet sense glossed as festival: a day or period of time set aside for feasting and celebration, the most sensible choice given that "Chinese New Year" does not appear as a WordNet concept. Annotating the IS-A relationship for the set was also important because the algorithm being evaluated was working on groups of head nouns, thereby potentially losing information pointing to a more specific concept reading. For example, the definition

(e) 钢管: steel tube, steel pipe

would be given to the algorithm as a group containing head nouns *tube* and *pipe*.

Once the test set was annotated, evaluation was done according to two paradigms: *selection* and *filtering*. In both paradigms we assume that for each entry in the test set, an annotator has correctly specified which WordNet senses are to be considered *correct*, and which are *incorrect*. An algorithm being tested against this set must identify, for each listed sense, whether that sense should be *included* for that item or whether it should be *excluded*. For example, the WordNet sense corresponding to "the murder or execution of someone" would be identified by an annotator as incorrect for (d), and so an algorithm marking it as "included" should be penalized.

For the *selection paradigm*, the goal is to identify WordNet senses to *include*. We can therefore define precision in that paradigm as

$$P_{\text{selection}} \quad = \quad \frac{\text{number of correctly included senses}}{\text{number of included senses}} \tag{10}$$

and recall as

$$R_{\text{selection}} \quad = \quad \frac{\text{number of correctly included senses}}{\text{number of correct senses}}. \tag{11}$$

These correspond directly to the use of precision and recall in information retrieval. Precision begins with the set of senses included by some method, and computes the proportion of these that are correct. Recall begins with the set of senses that *should* have been included, and computes the proportion of these that the method actually managed to choose.

Since the number of potential WordNet senses for an item can be quite large, an equally valid alternative to the selection paradigm is what I will call the *filtering paradigm*, according to which the goal is to identify WordNet senses to *exclude*. One can easily imagine this being the more relevant paradigm — for example, in a semi-automated setting where one wishes to reduce the burden of a user selecting among alternatives. In the filtering paradigm one can define filtering precision as

$$P_{\text{filtering}} \quad = \quad \frac{\text{number of correctly excluded senses}}{\text{number of excluded senses}} \tag{12}$$

and filtering recall as

$$R_{\text{filtering}} \quad = \quad \frac{\text{number of correctly excluded senses}}{\text{number of senses labeled incorrect}}. \tag{13}$$

In the filtering paradigm, precision begins with the set of senses that the method filtered out and computes the proportion that were *correctly* filtered out. And recall in filtering begins with the set of senses that *should* have been excluded (i.e. the incorrect ones) and computes the proportion of these that the method actually managed to exclude.

| | Sense Selection | | Sense Filtering | |
|---|---|---|---|---|
| | Precision (%) | Recall (%) | Precision (%) | Recall (%) |
| Random | 29.5 | 31.2 | 88.0 | 87.1 |
| Algorithm | 36.9 | 69.9 | 93.8 | 79.3 |
| Judge 2 | 54.8 | 55.6 | 91.9 | 91.7 |

Table 9: Evaluation using Judge 1 as the reference standard, considering items selected with confidence 3 and above.

| | | Judge 2 | | Algorithm | | Random | |
|---|---|---|---|---|---|---|---|
| | | Include | Exclude | Include | Exclude | Include | Exclude |
| Judge 1 | Include | 40 | 32 | 58 | 25 | 26 | 57 |
| | Exclude | 33 | 363 | 99 | 380 | 61 | 418 |

Table 10: Agreement and disagreement with Judge 1

Table 9 shows the precision/recall figures using the judgments of Judge 1, the native Chinese speaker, as a reference standard, considering only known items selected with confidence 3 and above.[18] The algorithm recorded all 100 items as known, and its confidence values were scaled linearly from continuous values in range [0,1] to discrete values from 1 to 5. The table shows the algorithm's results with its choice thresholded at confidence 3, and Figure 5 shows how recall and precision vary as the confidence threshold changes. As a lower bound for comparison, an algorithm was implemented that considered each word sense for each item, selecting that sense probabilistically (with complete confidence) in such a way as to make the average number of senses per item as close as possible to the average number of senses per item in the reference standard (1.3 senses). Figures for the random baseline are the average over 10 runs. Table 10 illustrates the choices underlying those figures; for example, there were 26 senses the random procedure chose to include that were also included by Judge 1.

The fact that Judge 2 has such low precision and recall for selection indicates that matching the choices of an independent judge is indeed a difficult task. This is unsurprising, given previous experience with the problem of selecting among WordNet's fine-grained senses (Resnik, 1998a; Resnik & Yarowsky, 1997). The results clearly show that the algorithm is better than the baseline, but also indicate that it is overgenerating senses, which hurts selection precision. In terms of filtering, when the algorithm chooses to filter out a sense it tends to do so reliably (filtering precision). However, its propensity toward overgeneration is reflected in its below-baseline performance on filtering recall; that is, the algorithm is choosing to allow in senses that it should be filtering out.

---

18. Judge 1, the native speaker of Chinese, identified 65 of the words as known to him; Judge 2 identified 69. This on-line dictionary was constructed from a large variety of lexical resources, and includes a great many uncommon words, archaic usages, regionalisms, and the like.

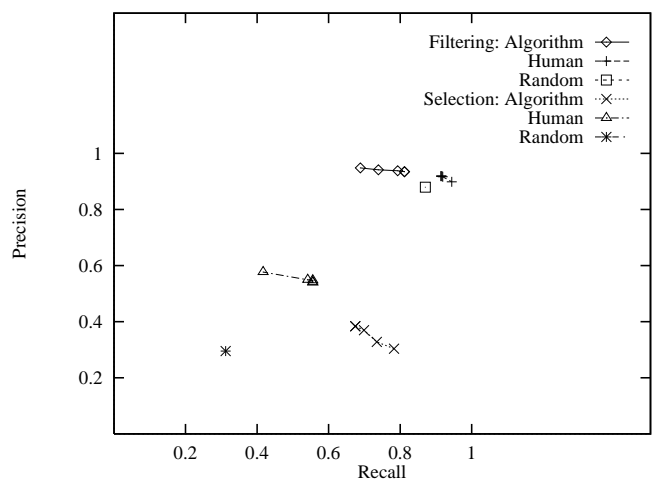

Figure 5: Precision/recall curves using Judge 1 as the reference standard, varying the confidence threshold

This pattern of results suggests that the best use of this algorithm at its present level of performance would be as a filter for a lexical acquisition process with a human in the loop, dividing candidate WordNet senses for dictionary entries according to higher and lower priority. For Chinese-English dictionary entries that serve as appropriate input to the algorithm (of which there are approximately 37000 in the CETA dictionary), if a WordNet sense is not selected by the algorithm with a confidence at least equal to 3 it should be demoted to the lower priority group in the presentation of alternatives, since the algorithm's choice to exclude a sense is correct approximately 93% of the time. Those senses that are selected by the algorithm are not *necessarily* to be included — the human judge is still needed to make the selection, since selection precision is low — but the algorithm tends to err on the side of caution, and so correct senses will be found in the higher priority group some 70% of the time.

### 5.3 Linking to WordNet from an English Dictionary/Thesaurus

The results on WordNet sense selection using a bilingual dictionary demonstrate that the algorithm of Figure 3 does a good job of assigning low scores to WordNet senses that should be filtered out, even if it should probably not be trusted to make categorical decisions. One application proposed as suitable, therefore, was helping to identify which senses should be filtered out within a semi-automated process of lexical acquisition. Here I describe a closely related, real-world application for which the algorithm has been deployed: adding pointers into WordNet from an on-line dictionary/thesaurus on the Web.

The context of this application is the Wordsmyth English Dictionary-Thesaurus (WEDT, `http://www.wordsmyth.net/`), an on-line educational dictionary affiliated with the ARTFL text database project (`http://humanities.uchicago.edu/ARTFL/`; Morrissey, 1993). It has been designed to be useful in educational contexts, and, as part of that design, it integrates a thesaurus within the structure of the dictionary. As illustrated in Figure 6,

**bar**
| | |
|---|---|
| SYL: | bar[1] |
| PRO: | <u>bar</u> |
| POS: | noun |
| DEF: | 1. a length of solid material, usu. rectangular or cylindrical: |
| EXA: | *a bar of soap;* |
| EXA: | *a candy bar;* |
| EXA: | *an iron bar.* |
| SYN: | rod (1), stick[1] (1,2,3) |
| SIM: | pole[1], shaft, stake[1], ingot, block, rail[1], railing, crowbar, jimmy, lever |
| DEF: | 2. anything that acts as a restraint or hindrance. |
| SYN: | block (10), hindrance (1), obstruction (1), impediment (1), obstacle, barrier (1,3), stop (5) |
| SIM: | barricade, blockade, deterrent, hurdle, curb, stumbling block, snag, jam[1], shoal[1], reef[1], sandbar |

⋮

Figure 6: Example from the Wordsmyth English Dictionary-Thesaurus (WEDT)

WEDT contains traditional dictionary information, such as part of speech, pronunciation, and definitional information, but in many cases also includes pointers to synonyms (SYN) or similar words (SIM). Within the on-line dictionary, these thesaurus items are hyperlinks — for example, stake[1] is a link to the first WEDT entry for *stake* — and parenthetical numbers refer to specific definitions within an entry.

The thesaurus-like grouping of similar words provides an opportunity to exploit the algorithm for disambiguating noun groupings by automatically linking WEDT entries to WordNet. The value in linking these two resources comes from their compatability, in that both have properties of both a thesaurus and a dictionary, as well as from their complementarity: beyond being an alternative source of definitional information and lists of synonyms, WordNet provides ordering of word senses by frequency, estimates of word familiarity, PART-OF relationships, and of course the overall taxonomic organization illustrated in Figures 1 and 2. Figure 7 shows how taxonomic information is presented using the WordNet Web server (`http://www.cogsci.princeton.edu/cgi-bin/webwn/`).

In a collaboration with WEDT and ARTFL, I have taken the noun entries from the WEDT dictionary and, for each grouping of similar words, added a set of experimental hyperlinks to WordNet entries on the WordNet Web server. Figure 8 shows how the experimental WordNet links (XWN) look to the WEDT user. Links to WordNet senses, such as $pole_1$, appear together with the confidence level assigned by the sense disambiguation algorithm; senses with confidence less than a threshold are not presented.[19] When an XWN hyperlink is selected by the user, WordNet taxonomic information for the selected sense appears in a parallel browser window, as in Figure 7.

From this window, the user has an entry point into the other capabilities of the WordNet web server. For example, one might choose to look at all the WordNet senses for *pole* as

---

19. The current threshold, 0.1, was chosen manually. It may be sub-optimal but I have found that it works well in practice.

```
Sense 1
pole
(a long (usually round) rod of wood or metal or plastic)
  => rod
     (a long thin implement made of metal or wood)
    => implement
       (a piece of equipment or tool used to effect an end)
      => instrumentality, instrumentation
         (an artifact (or system of artifacts) that is
         instrumental in accomplishing some end)
        => artifact, artefact
           (a man-made object)
         => object, physical object
            (a physical (tangible and visible) entity; ``it was
            full of rackets, balls and other objects'')
           => entity, something
              (anything having existence (living or nonliving))
```

Figure 7: WordNet entry (hypernyms) for pole[1]

**bar**

| | |
|---|---|
| SYL: | bar[1] |
| PRO: | <u>bar</u> |
| POS: | noun |
| DEF: | 1. a length of solid material, usu. rectangular or cylindrical: |
| EXA: | *a bar of soap;* |
| EXA: | *a candy bar;* |
| EXA: | *an iron bar.* |
| SYN: | rod (1), stick[1] (1,2,3) |
| SIM: | pole[1], shaft, stake[1], ingot, block, rail[1], railing, crowbar, jimmy, lever |
| XWN: | $pole_1$ *(0.82)* $ingot_1$ *(1.00)* $block_1$ *(0.16)* $rail_1$ *(0.39)* $railing_1$ *(1.00)* $crowbar_1$ *(1.00)* $jimmy_1$ *(1.00)* $lever_1$ *(0.67)* $lever2_2$ *(0.23)* $lever3$ *(0.15)* |
| DEF: | 2. anything that acts as a restraint or hindrance. |
| SYN: | block (10), hindrance (1), obstruction (1), impediment (1), obstacle, barrier (1,3), stop (5) |
| SIM: | barricade, blockade, deterrent, hurdle, curb, stumbling block, snag, jam[1], shoal[1], reef[1], sandbar |
| XWN: | $barricade_1$ *(1.00)* $barricade2$ *(1.00)* $blockade_1$ *(0.25)* $blockade_2$ *(0.75)* $deterrent_1$ *(1.00)* $hurdle_1$ *(0.50)* $hurdle_2$ *(0.43)* $curb_1$ *(0.56)* $curb_2$ *(0.56)* $curb_3$ *(0.29)* $curb_4$ *(0.44)* $stumbling\_block_1$ *(1.00)* $snag_1$ *(1.00)* $jam_1$ *(0.27)* $shoal_1$ *(0.23)* $shoal2$ *(0.91)* $reef_1$ *(1.00)* $sandbar_1$ *(1.00)* |

Figure 8: Example from WEDT with experimental WordNet links

1. pole – (a long (usually round) rod of wood or metal or plastic)

2. Pole – (a native or inhabitant of Poland)

3. pole – (one of two divergent or mutually exclusive opinions; "they are at opposite poles" or "they are poles apart")

4. perch, rod, pole – ((British) a linear measure of 16.5 feet)

5. perch, rod, pole – (a square rod of land)

6. pole, celestial pole – (one of two points of intersection of the Earth's axis and the celestial sphere)

7. pole – (one of two antipodal points where the Earth's axis of rotation intersects the Earth's surface)

8. terminal, pole – (a point on an electrical device (such as a battery) at which electric current enters or leaves)

9. pole – (a long fiberglass implement used for pole vaulting)

10. pole, magnetic pole – (one of the two ends of a magnet where the magnetism seems to be concentrated)

Figure 9: List of WordNet senses for *pole*

a noun, displayed as in Figure 9. Notice that if a user of WEDT had simply gone directly to the WordNet server to look up *pole*, the full list of 10 senses would have appeared with no indication of which are most potentially related to the WEDT dictionary entry under consideration. In contrast, the WEDT hyperlinks, introduced via the sense selection algorithm, filter out the majority of the irrelevant senses and provide the user a measure of confidence in selecting among those that remain.

Although no formal evaluation of the WEDT/WordNet connection has been attempted, the results of the bilingual dictionary experiment suggest that this application of word sense disambiguation — filtering out the least relevant senses, and then leaving the user in the loop — is a task for which the sense disambiguation algorithm is well suited. This is supported by user feedback on the XWN feature of WEDT, which has been favorable (Robert Parks, personal communication). The site has been growing in popularity, with a current estimate of 1000-1500 hits per day.

## 6. Related Work

There is an extensive literature on measuring similarity in general, and on word similarity in particular; for a classic paper see Tversky (1977). Recent work in information retrieval and computational linguistics has emphasized a distributional approach, in which words are represented as vectors in a space of features and similarity measures are defined in terms of those vectors; see Resnik (1998b) for discussion, and Lee (1997) for a good recent example. Common to the traditional and the distributional approaches is the idea that word or concept representations include explicit features, whether those features are specified in a knowledge-based fashion (e.g., *dog* might have features like MAMMAL, LOYAL) or defined in terms of distributional context (e.g., *dog* might have features like "observed within ±5 words of *howl*"). This representational assumption contrasts with the assumptions embodied in a taxonomic representation, where most often the IS-A relation stands between non-decomposed concepts. The two are not inconsistent, of course, since concepts in a taxonomy sometimes *can* be decomposed into explicit features, and the IS-A relation, as it is usually interpreted, implies inheritance of features whether they are explicit or implicit. In that respect, the traditional approach of counting edges can be viewed as a particularly simple approximation to a similarity measure based on counting feature differences, under the assumption that an edge exists to indicate a difference of at least one feature.

Information-theoretic concepts and techniques have, in recent years, emerged from the speech recognition community to find wide application in natural language processing; e.g., see Church and Mercer (1993). The information of an event is a fundamental notion in stochastic language modeling for speech recognition, where the contribution of a correct word prediction based on its conditional probability, p(word|context), is measured as the information conveyed by that prediction, $-\log$ p(word|context). This forms the basis for standard measures of language model performance, such as cross entropy. Frequency of shared and unshared features has also long been a factor in computing similarity over vector representations. The inverse document frequency (idf) for term weighting in information retrieval makes use of logarithmic scaling, and serves to identify terms that do not discriminate well among different documents, a concept very similar in spirit to the idea that such terms have low information content (Salton, 1989).

Although the counting of edges in IS-A taxonomies seems to be something many people have tried, there seem to be few published descriptions of attempts to directly evaluate the effectiveness of this method. A number of researchers have attempted to make use of conceptual distance in information retrieval. For example, Rada et al. (1989, 1989) and Lee et al. (1993) report experiments using conceptual distance, implemented using the edge-counting metric, as the basis for ranking documents by their similarity to a query. Sussna (1993) uses semantic relatedness measured with WordNet in word sense disambiguation, defining a measure of distance that weights different types of links and also explicitly takes depth in the taxonomy into account.

Following the original proposal to measure semantic similarity in a taxonomy using information content (Resnik, 1993b, 1993a), a number of related proposals have been explored. Leacock and Chodorow (1994) define a measure resembling information content, but using the normalized path length between the two concepts being compared rather than the probability of a subsuming concept. Specifically, they define

$$\text{wsim}_{\text{ndist}}(w_1, w_2) \;\;=\;\; -\log \left[ \frac{\min\limits_{c_1, c_2} \text{len}(c_1, c_2)}{(2 \times \text{MAX})} \right]. \tag{14}$$

(The notation above is the same as for Equation (5).) In addition to this definition, they also include several special cases, most notably to avoid infinite similarity when $c_1$ and $c_2$ are exact synonyms and thus have a path length of 0. Leacock and Chodorow have experimented with this measure and the information content measure described here in the context of word sense disambiguation, and found that they yield roughly similar results. Implementing their method and testing it on the task reported in Section 3, I found that it actually outperformed the information-based measure slightly on that data set; however, in a follow-up experiment using a different and larger set of noun pairs (100 items), the information-based measure performed significantly better (Table 11).

Analyzing the differences between the two studies is illuminating. In the follow-up experiment, I used netnews archives to gather highly frequent nouns within related topic areas (to ensure that similar noun pairs occurred) and then selected noun pairings at random (in order to avoid biasing the follow-up study in favor of either algorithm). There is, therefore, a predominance of low-similarity noun pairs in the test data. Looking at the distribution of ratings for the noun pairs, as given by the two measures, it is evident that the Leacock and Chodorow measure is overestimating semantic similarity for many of the predominantly non-similar pairs. This stands to reason since the measure is identical whenever the edge distance is identical, regardless of whether the pair is high or low in the taxonomy (e.g., the distance between *plant* and *animal* is the same as the distance between *white oak* and *red oak*). In contrast, the information-based measure is sensitive to the difference, and better at avoiding spuriously high similarity values for non-similar pairs. On a related note, the edge-counting measure used in the follow-up study was a variant that computes path length through a virtual top node, rather than asserting zero similarity between words with no path connecting them in the existing WordNet taxonomy, as was done previously. Using the data set in the follow-up study, the information-based measure, at $r = .6894$, does significantly better than either of the edge-counting variants ($r = .4101$ and $r = .2777$); but going back to the original Miller and Charles data, the virtual-top-node variant does significantly better than the assert-zero edge distance measure, with its correlation of $r = .7786$ approaching that of the measure based on information content. This comparison between the follow-up study and the original Miller and Charles data illustrates quite clearly how the utility of a similarity measure can depend upon the distribution of items given by the task.

| Similarity method | Correlation |
|---|---|
| Information content | $r = .6894$ |
| Leacock and Chodorow | $r = .4320$ |
| Edge-counting | $r = .4101$ |

Table 11: Summary of experimental results in follow-up study.

Lin (1997, 1998) has recently proposed an alternative information-theoretic similarity measure, derived from a set of basic assumptions about similarity in a style reminiscent of the way in which entropy/information itself has a formal definition derivable from a set of basic properties (Khinchin, 1957). Formally, Lin defines similarity in a taxonomy as:

$$\text{sim}_{\texttt{Lin}}(c_1, c_2) \;\; = \;\; \frac{2 \times \log \text{p}(\bigcap_i C_i)}{\log \text{p}(c_1) + \log \text{p}(c_2)} \tag{15}$$

where the $C_i$ are the "maximally specific superclasses" of both $c_1$ and $c_2$. Although the possibility of multiple inheritance makes the intersection $\bigcap_i C_i$ necessary in principle, multiple inheritance is in fact so rare in WordNet that in practice one computes Equation (15) separately for each common ancestor $C_i$, using $\text{p}(C_i)$ in the numerator, and then takes the maximum (Dekang Lin, p.c.). Other than the multiplicative constant of 2, therefore, Lin's method for determining similarity in a taxonomy is essentially the information-based similarity measure of Equation 1, but *normalized* by the combined information content of the two concepts assuming their independence. Put another way, Lin's measure is taking into account not only commonalities but differences between the items being compared, expressing both in information-theoretic terms.

| Similarity method | Correlation |
|---|---|
| Information content | $r = .7947$ |
| $\text{sim}_{\texttt{Wu\&Palmer}}$ | $r = .8027$ |
| $\text{sim}_{\texttt{Lin}}$ | $r = .8339$ |

Table 12: Summary of Lin's results comparing alternative similarity measures

Lin's measure is theoretically well motivated and elegantly derived. Moreover, Lin points out that his measure will by definition yield the same value for $\text{sim}_{\texttt{Lin}}(x, x)$ regardless of the identity of $x$ — unlike information content, which has been criticized on the grounds that the value of self-similarity depends on how specific a concept $x$ is, and that two non-identical items $x$ and $y$ can be rated more similar to each other than a third item $z$ is to itself (Richardson et al., 1994). From a cognitive perspective, however, similarity comparisons involving self-similarity ("Robins are similar to robins"), as well as subclass relationships ("Robins are similar to birds"), have themselves been criticized by psychologists as anomalous (Medin, Goldstone, & Gentner, 1993). Moreover, experimental evidence with human judgments suggests that not all identical objects are judged equally similar, consistent with the information-content measure proposed here but contrary to Lin's measure. For example, objects that are identical and complex, such as twins, can seem more similar to each other than objects that are identical and simple, such as two instances of a simple geometric shape (Goldstone, 1999; Tversky, 1977). It would appear, therefore, that insofar as fidelity to human judgments is relevant, further experimentation is needed to evaluate the competing predictions of alternative similarity measures.

Wu and Palmer (1994) propose a similarity measure that is based on edge distances, but related to Lin's measure in the way it takes into account the most specific node dominating $c_1$ and $c_2$, characterizing their commonalities, while normalizing in a way that accounts for their differences. Revising Wu and Palmer's notation slightly, their measure is:

$$\text{sim}_{\texttt{Wu\&Palmer}}(c_1, c_2) \;\; = \;\; \frac{2 \times \text{d}(c_3)}{\text{d}(c_1) + \text{d}(c_2)} \qquad (16)$$

where $c_3$ is the maximally specific superclass of $c_1$ and $c_2$, $\text{d}(c_3)$ is its depth, i.e. distance from the root of the taxonomy, and $\text{d}(c_1)$ and $\text{d}(c_2)$ are the depths of $c_1$ and $c_2$ on the path through $c_3$.

Lin (1998) repeats the experiment of Section 3 for the information content measure, $\text{sim}_{\texttt{Lin}}$, and $\text{sim}_{\texttt{Wu\&Palmer}}$, reporting the results that appear in Table 12. Lin uses a sense-tagged corpus to estimate frequencies, and smoothed probabilities rather than simple relative frequency. His results show a somewhat higher correlation for $\text{sim}_{\texttt{Lin}}$ than the other measures. Further experimentation is needed in order to assess the alternative measures, particularly with respect to their competing predictions and the variability of performance across data sets. What seems clear, however, is that all these measures perform better than the traditional edge-counting measure.

## 7. Conclusions

This article has presented a measure of semantic similarity in an IS-A taxonomy, based on the notion of information content. Experimental evaluation was performed using a large, independently constructed corpus, an independently constructed taxonomy, and previously existing and new human subject data, and the results suggest that the measure performs encouragingly well and can be significantly better than the traditional edge-counting approach. Semantic similarity, as measured using information content, was shown to be useful in resolving cases of two pervasive kinds of linguistic ambiguity. In resolving coordination ambiguity, the measure was employed to capture the intuition that similarity of meaning is one indicator that two words are being conjoined; suggestive results of a first experiment were bolstered by unequivocal results in a second study, demonstrating significant improvements over a disambiguation strategy based only on syntactic agreement. In resolving word sense ambiguity, the semantic similarity measure was used to assign confidence values to word senses of nouns within thesaurus-like groupings. A formal evaluation provided evidence that the technique can produce useful results but is better suited for semi-automated sense filtering than categorical sense selection. Application of the technique to a dictionary/thesaurus on the World Wide Web provides a demonstration of the method in action in a real-world setting.

## Acknowledgements

Sections 1-3 of this article comprise a revised and extended version of Resnik (1995). Section 4 describes previously presented algorithms and data (Resnik, 1993b, 1993a), extended by further discussion and analysis. Section 5 summarizes an algorithm described in Resnik (1998a), and then extends previous results by presenting new applications of the algorithm, with Section 5.2 containing a formal evaluation in a new setting and Section 5.3 giving a real-world illustration where the approach has been put into practice. Section 6 adds a substantial discussion of related work by other authors that has taken place since the information-based similarity measure was originally proposed.

Parts of this research were done at the University of Pennsylvania with the partial support of an IBM Graduate Fellowship and grants ARO DAAL 03-89-C-0031, DARPA N00014-90-J-1863, NSF IRI 90-16592, and Ben Franklin 91S.3078C-1; parts of this research were also done at Sun Microsystems Laboratories in Chelmsford, Massachusetts; and parts of this work were supported at the University of Maryland by Department of Defense contract MDA90496C1250, DARPA/ITO Contract N66001-97-C-8540, Army Research Laboratory contract DAAL03-91-C-0034 through Battelle, and a research grant from Sun Microsystems Laboratories. The author gratefully acknowledges the comments of three anonymous JAIR reviewers and helpful discussions with John Kovarik, Claudia Leacock, Dekang Lin, Johanna Moore, Mari Broman Olsen, and Jin Tong, as well as comments and criticism received during various presentations of this work.